%% file: Main.tex
\documentclass{article}

    \PassOptionsToPackage{numbers, compress}{natbib}

 \usepackage[main, final]{neurips_2026}
\usepackage{graphicx}
\usepackage{wrapfig}
\usepackage{booktabs} 
\usepackage{enumitem}

\usepackage[utf8]{inputenc} 
\usepackage[T1]{fontenc}    
\usepackage{hyperref}       
\usepackage{url}            
\usepackage{booktabs}       
\usepackage{amsfonts}       
\usepackage{nicefrac}       
\usepackage{microtype}      
\usepackage{xcolor}         
\usepackage{amsmath}
\usepackage{multirow}

\title{UxSID: Semantic-Aware User Interests Modeling for Ultra-Long Sequence}

\author{%
  \textbf{Hongwei Zhang}\thanks{Equal contribution.\ \textsuperscript{\dag}Corresponding author.\ \textsuperscript{\ddag}Work done during an internship at Kuaishou Technology.}%
  \ \ \textsuperscript{, \ddag},
  \textbf{Qiqiang Zhong}\footnotemark[1],
  \textbf{Jiangxia Cao}\footnotemark[1],
  \textbf{Junfeng Shu}\textsuperscript{\dag}, 
  \textbf{Yiyang Lv}\textsuperscript{\ddag},\\
  \textbf{Huanjie Wang}\textsuperscript{\ddag},
  \textbf{Liwei Guan}\textsuperscript{\ddag},
  \textbf{Honggang Chen},
  \textbf{Jing Yao},
  \textbf{Tianlu Xie},\\
  \textbf{Chi Lu},
  \textbf{Yiyu Wang},
  \textbf{Mingxing Wen},
  \textbf{Zhaojie Liu},
  \textbf{Wenjin Wu},
  \textbf{Han Li} \\
  Kuaishou Technology \\
    \texttt{\{zhanghongwei08,lifan11,caojiangxia,shujunfeng,lvyiyang,}\\
    \texttt{wanghuanjie,guanliwei\}@kuaishou.com,}\\
    \texttt{yuanhong.chg@gmail.com,}\\
    \texttt{\{yaojing03,xietianlu,luchi,liubin21,wenmingxing,}\\
    \texttt{zhaotianxing,wuwenjin,lihan08\}@kuaishou.com}%
}

\begin{document}

\maketitle

\input{Abstract}

\input{Introduction}
\input{Related_Work}
\input{Methods}
\input{Experiments}
\input{Conclusion}
\newpage
{
\small 
\bibliographystyle{unsrtnat}
\bibliography{NIPS26.bib}
}
\input{Appendix}

\end{document}

%% file: Abstract.tex
\begin{abstract}

Modeling ultra-long user behavior sequences is an essential task for capturing evolving preferences in modern recommender systems, and this research direction has contributed solid gains in the past several years.
However, recommendation systems always serve enormous traffic, and extending user sequences sharply increases computational cost, which creates a difficult trade-off between efficiency and effectiveness.
To scale to longer user sequences while keeping lightweight serving computation, existing works can be divided into two paradigms: (1) \textit{Search-based Top-$K$ selection}, which constructs an \textbf{item-specific} subsequence for each candidate item to avoid facing the ultra-long sequence directly; and (2) \textit{pre-trained user-interest compression}, which maps an ultra-long user sequence into a small group of \textbf{item-agnostic} user interest memories so that the online model can perceive user long-term interests via this highly compressed dense memory.
Besides the two technical routes (totally item-specific or item-agnostic), we argue that there exists an intermediate path not well explored: preserving partial relevance between the user sequence and the target item, while exposing only limited signals to guide the direction of interest compression.
This design does not pursue item-specific user interest compression, but seeks semantic-group shared general user interest memory according to item attributes, where semantically similar items share the same compressed user interest memory.
Motivated by this, we propose \textbf{UxSID}, a novel framework that bridges this gap by facilitating a target item semantic-aware interaction between \textbf{U}ser histories and candidate \textbf{S}emantic \textbf{ID}s (SIDs).
Specifically, UxSID employs a dual-level attention strategy: it first extracts  item-agnostic user interests from raw sequences, and then performs a semantic-specific query over global behaviors and those agnostic interests to generate semantic-specific preferences.
By adopting this end-to-end architecture, UxSID generates offline embeddings that balance computational parsimony with target items' semantic awareness, while strictly preserving parity with online inference in constant time.
Extensive public benchmarks and a large-scale A/B test demonstrate that UxSID achieves state-of-the-art performance, driving a 0.337\% revenue lift in advertising.

\end{abstract}

%% file: Introduction.tex
\section{Introduction}
\label{Introduction}



Massive industrial platforms like TikTok, Instagram Reels, and Kuaishou attract vast user bases and host millions of creators producing images, short videos, and live streams.
To connect users and the platform's items, a powerful recommendation system (RecSys) is necessary to distribute the right items to the right users.
Within this recommendation and consumption cycle, hundreds of millions of users produce unprecedented volumes of daily behavioral data; for instance, active users often interact with as many as 10,000 items per week.
%
%
As a result, Ultra-Long Sequence Modeling (ULSM) is paramount for modern recommender systems; it captures comprehensive preference evolution over extended horizons, bridging the gap between noisy short-term signals and underlying long-term user signals.
By uncovering associations overlooked by truncated models, ULSM serves as a key driver for enhancing user engagement, conversion rates, and long-term loyalty \cite{ULRM,mimn,mit}.

\begin{figure}
    \label{fig:motivation}
    \centering
    \includegraphics[width=1\linewidth]{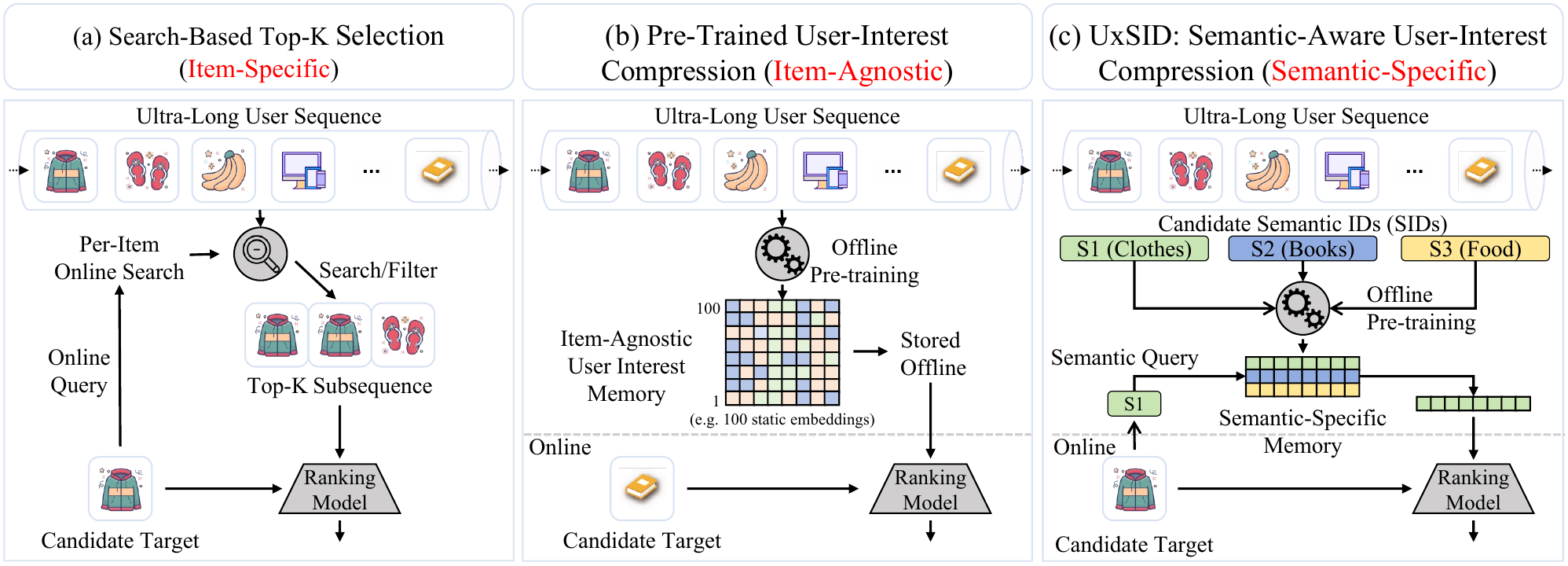}
    \caption{\textbf{Comparison of different paradigms for ULSM.} (a) \textbf{Item-specific Search}: Online filtering for each candidate, incurring high computational cost. (b) \textbf{Item-agnostic Compression}: Offline distillation into static memories, lacking target-specificity. (c) \textbf{UxSID}: A semantic-specific path that shares compressed interest memories among items with the same SIDs.}
\end{figure}

Despite its importance, scaling sequence length introduces heavy computational burdens on industrial recommender systems that must handle massive real-time traffic.
For instance, standard attention-based models \cite{attention} such as DIN \cite{din}, DIEN \cite{dien}, and TransAct \cite{transact} are inherently constrained by the computational complexity; consequently, many systems only afford limited subsequence lengths of $n=100$.
As a result, many RecSys researchers are focusing on developing high-efficiency methods to scale user sequences with lightweight computation.
Generally, there are two major ULSM technical roadmaps that have been well explored in recent years (a more comprehensive discussion of \textbf{Related Work} is deferred to Appendix \ref{Related Work}):
\begin{itemize}[leftmargin=*]
    \item Item-specific Top-$K$ subsequence selection (Figure~\ref{fig:motivation}(a)). Methods such as SIM \cite{SIM} and TWIN \cite{twin} utilize a Global Search Unit (GSU) to retrieve a Top-$K$ subset of behaviors relevant to the target item prior to fine-grained modeling. Despite their efficiency, search-based methods suffer from inherent selection bias: whether via discrete hard matching or embedding-based soft retrieval, the filtering process is fundamentally constrained by the expressiveness of the pre-defined key space and stringent selection quotas.
    \item Item-agnostic compressed user interest memories (Figure~\ref{fig:motivation}(b)). Compression-based approaches, such as MIMN \cite{mimn}, LURM \cite{lurm} and C-Former \cite{C-former}, aim to distill extensive histories into compact latent representations. However, these embeddings are typically target-agnostic, as they attempt to condense a user’s diverse, evolving interests into static, user-centric embeddings \cite{C-former,pinnerformer}. Consequently, these methods inevitably introduce irrelevant noise \cite{review-noise} and lack the target specificity required to disentangle a user's current intent from their vast historical background.
\end{itemize}


\begin{figure}
    \centering
    \includegraphics[width=1\linewidth]{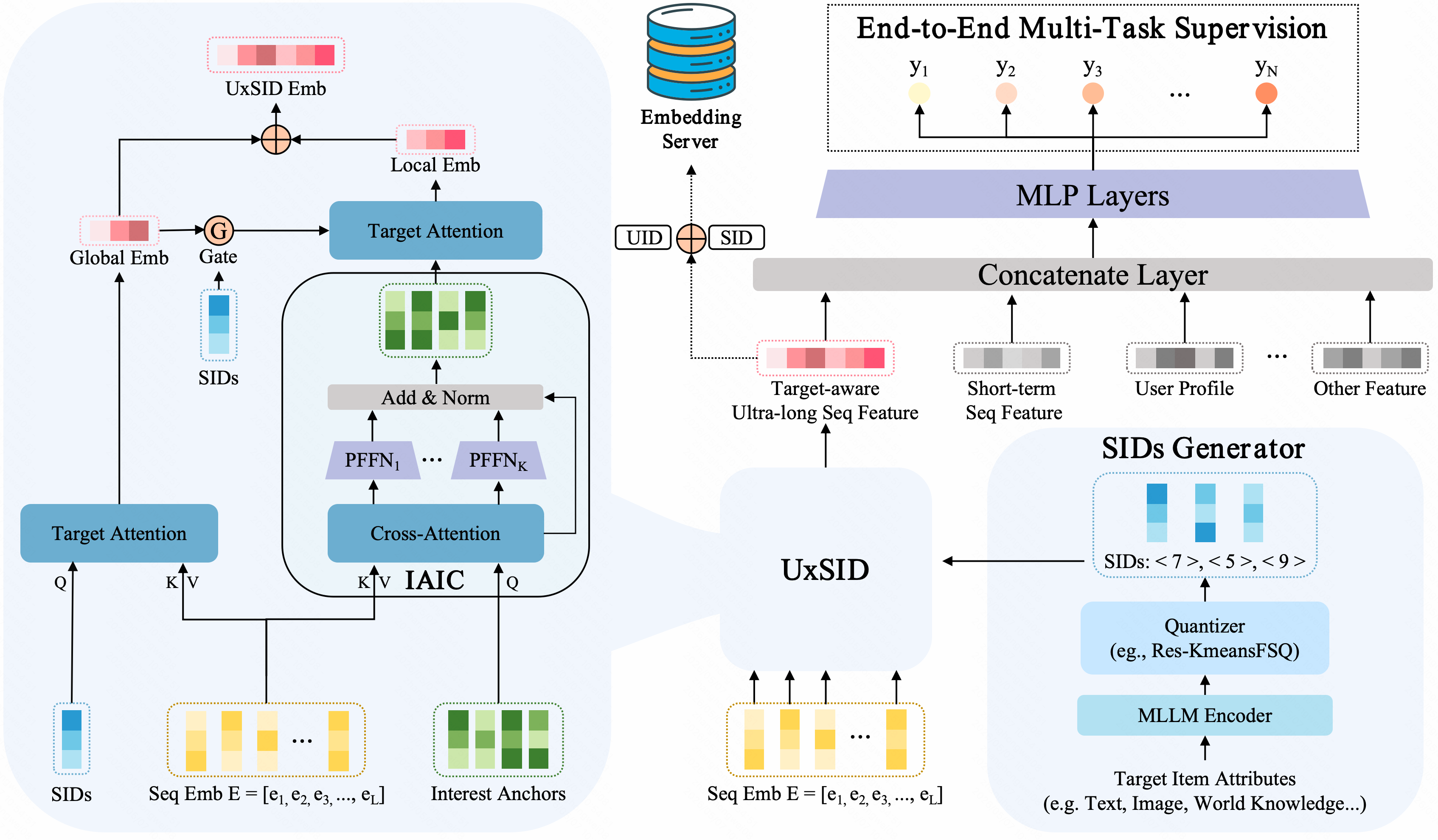}
    \caption{The architecture of UxSID primarily comprises three components: a target SIDs Generator that quantizes comprehensive attributes into semantically rich discrete IDs; a semantic-aware compression network featuring interest compression and hierarchical attention for sequence modeling; and an end-to-end multi-task supervision framework to ensure online-offline consistency.}
    \label{fig:overview}
\end{figure}

Besides the two technical routes (fully item-specific or entirely item-agnostic with target candidates), we argue that there exists an intermediate path not well explored: preserving partial relevance between the user sequence and the target item, while exposing only limited signals to guide the direction of interest compression. This design does not pursue ``item-specific'' user interest compression, but seeks semantic-group shared general user interest memory according to item attributes, where semantically similar items share the same compressed user interest memory, as shown in Figure~\ref{fig:motivation}(c).

To operationalize this intermediate path, we propose \textbf{UxSID} (Figure \ref{fig:overview}), a target semantic-aware compression framework that leverages Semantic IDs (SIDs) as the principled medium for interest distillation. Our core intuition is that SIDs, derived from deep quantization (e.g., RQ-VAE \cite{tiger}), act as high-density semantic probes that naturally align with user interest clusters. While traditional IDs are discrete symbols, SIDs represent synthesized semantic clusters derived from multimodal content \cite{onemall,plum}, collaborative signals \cite{ye2025das,pit,dos}, and world knowledge \cite{qarmv2}. By utilizing SIDs to guide the compression process, UxSID can precisely extract target-relevant signals from expansive histories, rendering candidate-aware modeling both feasible and efficient for large-scale industrial deployment.

Specifically, we propose a hierarchical compression-and-probe strategy to ensure both global historical coverage and semantic-specific precision:

\begin{itemize}[leftmargin=*]
\item \textbf{Item-Agnostic Interest Compression (IAIC).} This module condenses heterogeneous histories into a compact set of learnable interest anchors. By employing Per-token Feed-Forward Networks (PFFNs) and an orthogonality constraint, IAIC ensures each distilled interest captures a distinct facet of multifaceted preferences, effectively preserving representation diversity.
\item \textbf{Hierarchical Semantic Probing.} It first extracts fine-grained global signals from the ultra-long sequence, then employs a gated semantic query to dynamically integrate information from the compressed interests. This process selectively amplifies target-relevant interests while filtering noise, ensuring the final representation encapsulates both immediate intent and long-term dependencies.
\end{itemize}

By integrating this strategy, UxSID generates compressed representations that maintain strict alignment between offline training and online inference. This architecture effectively overcomes the inherent selection bias of traditional paradigms, achieving target semantic-level resolution within expansive interaction histories. Our primary contributions are summarized as follows:
\begin{itemize}[leftmargin=*]
\item \textbf{Target Semantic-Aware Behavior Compression.} To the best of our knowledge, UxSID is the first to enable SID-based semantic-specific compression in ULSM. By leveraging SIDs as queries, we shift the paradigm from static, user-centric distillation to target-aware interest compression.

\item \textbf{Scalable UxSID Architecture.} We design a dual-stage framework featuring IAIC and Hierarchical Target Attention. This architecture ensures constant-time inference and stable resource consumption even as sequences scale to 10k interactions, effectively translating long-term dependencies into consistent ranking gains.

\item \textbf{SOTA Performance and Industrial Impact.} We provide comprehensive validation on both public benchmarks and large-scale deployments at Kuaishou. UxSID consistently outperforms SOTA baselines and has delivered a 0.337\% revenue increase in production, demonstrating its robust scalability and significant business value under stringent industrial latency constraints.
\end{itemize}

\textbf{Paper Overview.} Section \ref{sec:methods} elaborates on the proposed methodology of UxSID and its industrial application pipeline. Section \ref{sec:settings} presents extensive experimental results, spanning offline evaluations, a online A/B test, rigorous ablation studies, and case studies. Finally, Section \ref{Conclusion} concludes the paper.

%% file: Methods.tex
\section{Methodology}
\label{sec:methods}
In this section, we present the details of UxSID, a semantic-aware compression framework for ULSM. UxSID utilizes a hierarchical attention architecture to perform end-to-end interest compression and recommendation, ensuring high-fidelity interest perception.

\textbf{Architecture Overview.} The UxSID architecture comprises three core modules: (1) \textbf{SIDs Generation}, which leverages a Multimodal Large Language Model (MLLM) to transform heterogeneous item content into discrete semantic codes; (2) \textbf{Item-Agnostic Interest Compression}, which filters raw trajectories into structured interest anchors; and (3) \textbf{Hierarchical Semantic Probing}, which employs a gated dual-stage attention mechanism to resolve target-specific intent.

\textbf{Semantic IDs Generation.} To empower the model with a semantic probe that transcends literal ID matching, we generate SIDs via a reasoning-based alignment mechanism \cite{qarmv2}. Given an item $i$ with multimodal attributes (e.g., video frames and textual descriptions), we first employ an MLLM encoder to project it into a continuous business-aligned semantic space:
\begin{equation}
\mathbf{z}_i = \text{Enc}_{\text{MLLM}}(\text{Attributes}_i)
\end{equation}
where $\mathbf{z}_i \in \mathbb{R}^d$. To ensure computational and storage efficiency, we apply a Res-KmeansFSQ hybrid quantization method \cite{FSQ}. This process decomposes $\mathbf{z}_i$ into $M$ hierarchical levels:
\begin{equation}
\mathbf{z}_i \approx \sum_{m=1}^{M} \mathcal{C}_m(k_m), \quad k_m = \arg\min_{j} \| \mathbf{r}_{m-1} - \mathbf{c}_{m,j} \|_2
\end{equation}
where $\mathcal{C}_m$ represents the codebook at the $m$-th level, and $\mathbf{c}_{m,j} \in \mathbb{R}^d$ is the $j$-th codeword within that codebook. The vector $\mathbf{r}_m = \mathbf{z}_i - \sum_{l=1}^{m} \mathcal{C}_l(k_l)$ denotes the quantization residual at level $m$, with the initial residual defined as $\mathbf{r}_0 = \mathbf{z}_i$. Through this approach, the item is represented by a sequence of SIDs $(k_1, k_2, \dots, k_M)$. In our industrial deployment, we primarily utilize the first-layer code $k_1$ as the target SIDs ($c_{target}$) to strike a balance between semantic granularity and inference latency.

\textbf{Item-Agnostic Interest Compression.}
The IAIC block aims to condense the raw interaction sequence $\mathcal{B} = [b_1, b_2, \dots, b_L]$ into a compact set of $K$ interest anchors $\mathbf{P} \in \mathbb{R}^{K \times d}$ ($K \ll L$).

\textit{Interest Anchor Compression.} We first transform each item $i_t$ into a $d$-dimensional representation through an embedding lookup layer, resulting in the matrix $\mathbf{E} = [\mathbf{e}_1, \mathbf{e}_2, \dots, \mathbf{e}_L] \in \mathbb{R}^{L \times d}$. To extract representative signals, we define a set of learnable interest anchors $\mathbf{Q}_{anc} \in \mathbb{R}^{K \times d}$ that serve as interest queries to aggregate salient features via a cross-attention mechanism:
\begin{equation}
\mathbf{H} = \text{Softmax}\left( \frac{(\mathbf{Q}_{anc}\mathbf{W}^Q)(\mathbf{E}\mathbf{W}^K)^\top}{\sqrt{d}} \right) (\mathbf{E}\mathbf{W}^V)
\end{equation}
where $\mathbf{H} = [\mathbf{h}_1, \dots, \mathbf{h}_K]^\top \in \mathbb{R}^{K \times d}$ represents the compressed interest features, and $\mathbf{W}^Q, \mathbf{W}^K, \mathbf{W}^V \in \mathbb{R}^{d \times d}$ are learnable projection matrices. Besides, to enhance the independent representation of each interest anchor, we employ PFFNs. Specifically, for each anchor $\mathbf{h}_k$, we apply an anchor-specific sub-network followed by a residual connection and layer normalization:
\begin{equation}
\mathbf{p}_k = \text{LayerNorm}\left( \mathbf{h}_k + \sigma(\mathbf{h}_k \mathbf{W}_1^{(k)} + \mathbf{b}_1^{(k)})\mathbf{W}_2^{(k)} + \mathbf{b}_2^{(k)} \right)
\end{equation}
where $\mathbf{W}_1^{(k)}, \mathbf{W}_2^{(k)}$ and $\mathbf{b}_1^{(k)}, \mathbf{b}_2^{(k)}$ are the learnable parameters specific to the $k$-th interest anchor, and $\sigma(\cdot)$ is the sigmoid function. This architecture ensures that the transformation is applied to each interest anchor independently, allowing them to be refined within their own semantic sub-spaces. The final set of compressed anchors is denoted as $\mathbf{P} = [\mathbf{p}_1, \dots, \mathbf{p}_K]^\top \in \mathbb{R}^{K \times d}$. 

\textit{Diversity and Orthogonality Constraint.} To ensure the interest anchors capture multifaceted preferences and avoid degenerate convergence toward a singular interest, we introduce a normalized Orthogonality Loss $\mathcal{L}_{ortho}$. By incorporating the squared $L_2$ norm of $\mathbf{P}$ in the denominator, we enforce structural independence between anchors while maintaining stability across varying scales:
\begin{equation}
\mathcal{L}_{ortho} = \left\| \frac{\mathbf{P}\mathbf{P}^T}{\|\mathbf{P}\|_2^2} - \mathbf{I} \right\|_F
\end{equation}
where $\mathbf{I}$ is the identity matrix and $\| \cdot \|_F$ denotes the Frobenius norm. This constraint minimizes the redundancy between interest anchors, ensuring that each interest in the latent representation represents a distinct and well-separated facet of the user's long-term behavioral trajectory.

\textbf{Hierarchical Semantic Probing.}
Unlike static methods, UxSID utilizes the target SIDs $c_{target}$ as an active probe through a hierarchical dual-stage attention mechanism, refined by a gating module.

\textit{Explicit Semantic Probing.} The first stage extracts fine-grained semantic signals by directly attending to the raw behavior sequence $\mathbf{E}$ using the target SIDs. This process captures global correlations between the target item and the user's expansive history:
\begin{equation}
\mathbf{e}_{global} = \text{Softmax}\left( \frac{(c_{target}\mathbf{W}_{g}^Q)(\mathbf{E}\mathbf{W}_{g}^K)^\top}{\sqrt{d}} \right) (\mathbf{E}\mathbf{W}_{g}^V)
\end{equation}
where $\mathbf{e}_{global} \in \mathbb{R}^d$ represents the global interest response.


\textit{Gated Latent Probing.} To refine interest resolution, the second stage conditions the probing process on the global context via a gating vector $\mathbf{g}_{ctx} \in \mathbb{R}^d$:
\begin{equation}
\mathbf{g}_{ctx} = \sigma\left( \text{GatedNet}(\mathbf{e}_{global}) \right)
\end{equation}
where $\text{GatedNet}(\cdot)$ is a two-layer MLP and $\sigma(\cdot)$ is the sigmoid function. This gating vector acts as a latent mask, modulating the target SID embedding $c_{target}$ to produce a refined query $\mathbf{q}_{ref}$:
\begin{equation}
\mathbf{q}_{ref} = c_{target} \odot \mathbf{g}_{ctx}
\end{equation}
By applying this Hadamard product, we align target semantic attributes with the user's global behaviors. Finally, the localized intent $\mathbf{e}_{local}$ is extracted from the interest anchors $\mathbf{P}$ via:
\begin{equation}
\mathbf{e}_{local} = \text{Softmax}\left( \frac{(\mathbf{q}_{ref}\mathbf{W}_{l}^Q)(\mathbf{P}\mathbf{W}_{l}^K)^\top}{\sqrt{d}} \right) (\mathbf{P}\mathbf{W}_{l}^V)
\end{equation}
where $\mathbf{W}_{l}^Q, \mathbf{W}_{l}^K, \mathbf{W}_{l}^V \in \mathbb{R}^{d \times d}$ are learnable parameters.

The final target-aware representation is obtained by concatenating the dual-stage outputs: $\mathbf{E}^{\text{UxSID}} = [\mathbf{e}_{global} ; \mathbf{e}_{local}]$. This hierarchical probing ensures that UxSID can simultaneously perceive both the broad historical context and the specific latent interest peaks relevant to the real-time recommendation.

\textbf{Model Training and Loss Function.} The precomputed $\mathbf{E}^{\text{UxSID}}$ is integrated with target features $\mathbf{E}^t$, user profiles $\mathbf{E}^u$, context $\mathbf{E}^c$, and short-term behaviors $\mathbf{E}^{\text{short}}$. The prediction $p(x)$ is formalized as:
\begin{equation}
p(x) = \sigma\left(\text{MLP}\left(\mathbf{E}^t; \mathbf{E}^u; \mathbf{E}^c; \mathbf{E}^{\text{short}}; \mathbf{E}^{\text{UxSID}} \mid x\right)\right)
\end{equation}
The model is optimized end-to-end via a joint loss function:
\begin{equation}
\mathcal{L} = -\frac{1}{N} \sum_{n=1}^{N} [y_n \log(p(x_n)) + (1-y_n) \log(1-p(x_n))] + \lambda \mathcal{L}_{ortho}
\end{equation}
where the first term is the Binary Cross-Entropy loss for the recommendation task.

\textbf{Serving in Production.} The offline pipeline ensures the final representation $\mathbf{E}^{\text{UxSID}}$ is target-aware and task-aligned. After training, $\mathbf{E}^{\text{UxSID}}$ is precomputed and cached in Embedding Server (ES). The storage key is generated via a bitwise concatenation of user ID ($UID$) and target item SID ($SID$):
\begin{equation}
\text{Key} = \text{Hash}(UID \oplus SID), \quad \text{Value} = \mathbf{E}^{\text{UxSID}}
\end{equation}
In the real-time serving stage, the system performs an $O(1)$ point lookup to retrieve $\mathbf{E}^{\text{UxSID}}$ from ES. The strong representational capacity of SIDs allows for a limited and manageable set of unique IDs per user; thus, the total storage footprint remains well within industrial feasibility (see Appendix \ref{appendix:a}). The compressed embeddings then function as target-aware user history keys/values, which interact with the target item query (augmented with side information) via a lightweight attention mechanism for final ranking, thereby effectively meeting stringent industrial latency requirements.

%% file: Experiments.tex
\section{Experiments}
\label{sec:settings}




\textbf{Datasets.} The effectiveness of UxSID is evaluated on two public benchmarks, XLong \cite{XLong} and KuaiRec-Big \cite{kuairec}, alongside a large-scale industrial dataset. These public benchmarks were selected because they provide paired content features, which are indispensable for the training and semantic alignment of SIDs. Detailed statistics and preprocessing procedures are provided in Appendix \ref{app:datasets}.

\textbf{Baselines.} To evaluate the effectiveness of UxSID, we compare it against several competitive baselines (which are detailed in Appendix \ref{app:baselines}): DIN \cite{din}, SIM \cite{SIM}, ETA \cite{eta}, SDIM \cite{sdim}, MIRRN \cite{MIRRN}, TWIN \cite{twin}, C-Former \cite{C-former}. All baselines share the bottom-level layers, varying only in ULSM.

\textbf{Metrics.} We evaluate model performance using AUC, UAUC, and WUAUC to measure overall and user-level ranking performance, referencing studies like \cite{twin,qarmv2,wide&deep}. Furthermore, we introduce Interest Recall@$K$ (Int.R@K) in ablation and case analysis studies to quantify semantic activation precision, defined as the proportion of the Top-$K$ behaviors (retrieved via attention score during explicit semantic probing) that share the same first-layer SIDs or category (tag) with the target item.

\textbf{Implementation Details.} We followed the experimental settings of \cite{C-former,MIRRN}, including data partitioning and hyperparameter configurations. Baseline evaluations use sequence lengths of 2k for KuaiRec-Big and 1k for XLong, except for DIN which uses 100. The GSU retrieval quantity is also set to 100. The prediction heads of all models consist of an MLP with $\{200,80,2\}$ hidden size, and the sparse embedding dimension is fixed at 16. Notably, main evaluations are conducted with 1k length for industrial dataset, while scalability up to 10k is specifically analyzed in Section \ref{sec:efficiency_scaling}. All public experiments are trained using the Adam optimizer with a batch size of 256 and a learning rate of 0.001. Specifically for UxSID, the gating network is configured as $\{16, 16\}$ with an activation layer, while each FFN in the PFFNs follows a $\{16, 32, 16\}$ structure. We adopted LETTER \cite{letter} to construct the codebook and train the SIDs with a shape of $\{256, 256, 256, 256\}$ for public datasets. Ultimately, the first-layer SIDs were utilized, and the number of IAIC anchors is 16.

\subsection{Overall Performance}

\begin{wraptable}{R}{0.5\textwidth}
\centering
\small
\caption{AUC Performance comparison with baselines. The best results are in bold. ‘-’ indicates category unavailability.}
\label{tab:public_results}
\begin{tabular}{lcc}
\toprule
\textbf{Models} & \textbf{XLong} & \textbf{KuaiRec-Big} \\
\midrule
DIN                      & 0.7889      & 0.8181        \\
SIM-Hard               & --          &  0.8201    \\
SIM-Soft                & 0.7971      & 0.8279        \\
ETA                      & 0.7910    & 0.8231          \\
SDIM                     & 0.7915     & 0.8209          \\
MIRRN                    & 0.7926     & 0.8217         \\
TWIN                     & 0.8154        & 0.8269      \\
C-Former               & 0.8135    & 0.8276          \\
\midrule
\textbf{UxSID} & \textbf{0.8408} & \textbf{0.8348}    \\
\bottomrule
\end{tabular}
\end{wraptable}

\textbf{Performance on Public Datasets.} We first conduct experiments on two widely-recognized public benchmarks: XLong and KuaiRec-Big. As illustrated in Table \ref{tab:public_results}, UxSID consistently achieves SOTA performance, outperforming all baseline paradigms. Furthermore, we show in Table \ref{tab:detailed_seed_results} (Appendix \ref{app:Stability Analysis}) that the standard deviation in the metrics for UxSID is insignificant.

Specifically, on XLong, UxSID achieves 0.8408 AUC, surpassing the strongest search-based baseline (TWIN) and the advanced compression model (C-Former). The gain over C-Former is noteworthy: while C-Former leverages learnable anchors for clustering, it remains inherently target-agnostic during compression. In contrast, UxSID utilizes SIDs as semantic queries to activate target-specific interests, demonstrating that target-aware perception is critical for resolving fine-grained preferences in ULSM.

\textbf{Performance on Large-Scale Industrial Datasets.} We further evaluate UxSID on a massive industrial dataset from Kuaishou, where a 0.1\% improvement represents a significant milestone.

As reported in Table \ref{tab:main_results}, UxSID achieves a CTCVR AUC of 0.8626, significantly outperforming the strongest industrial baselines, including SIM-Soft (+0.18\%) and TWIN (+0.17\%). This superior performance is primarily attributed to our IAIC block and the hierarchical probe architecture. While SIM and TWIN possess certain item-specific capabilities via retrieval or attention, UxSID further extends this advantage by leveraging the high-density semantics of SIDs to navigate the expansive interest landscape. Unlike traditional heuristic-based filtering which leads to information loss, UxSID’s superior structure maintains global interest-capturing capabilities by precisely bridging the gap between dense user histories and target semantic intents.

\begin{table}[ht]
\centering
\small
\caption{Performance comparison between UxSID and other baseline models.}
\label{tab:main_results}
\begin{tabular}{lcccccc}
\toprule
\multirow{2}{*}{\textbf{Models}} & \multicolumn{3}{c}{\textbf{CTR}} & \multicolumn{3}{c}{\textbf{CTCVR}} \\
\cmidrule(lr){2-4} \cmidrule(lr){5-7}
 & AUC & UAUC & WUAUC & AUC & UAUC & WUAUC \\
\midrule
SIM-Hard & 0.8698 & 0.6042 & 0.6063 & 0.8599 & 0.6161 & 0.6221 \\
SIM-Soft & 0.8711 & 0.6084 & 0.6099 & 0.8608 & 0.6228 & 0.6307 \\
TWIN     & 0.8712 & 0.6093 & 0.6104 & 0.8609 & 0.6232 & 0.6310 \\
\midrule
\textbf{UxSID (Ours)} & \textbf{0.8728} & \textbf{0.6125} & \textbf{0.6161} & \textbf{0.8626} & \textbf{0.6269} & \textbf{0.6350} \\
\bottomrule
\end{tabular}
\end{table}

\begin{wraptable}{R}{0.5\textwidth}
\centering
\small
\caption{Online A/B results.}
\label{tab:online_ab_test}
\begin{tabular}{l|ccc}
\toprule
\multirow{2}{*}{\textbf{Scenarios}} & \multicolumn{3}{c}{\textbf{Advertising Metrics}} \\
\cmidrule(lr){2-4}
 & Exposure & Cost & Revenue \\
\midrule
Advertising & +0.111\% & +0.231\% & +0.337\% \\
\bottomrule
\end{tabular}
\end{wraptable}

\textbf{Performance in Online A/B Testing.} We deployed UxSID on Kuaishou's short-video advertising platform, where a week-long online A/B test (Table \ref{tab:online_ab_test}) demonstrated its effectiveness through significant performance gains.

The significant divergence between Revenue (+0.337\%) and Exposure (+0.111\%) is particularly revealing, highlighting a shift towards high-precision conversion. These results demonstrate that, through semantic-specific compression and storage, UxSID renders end-to-end ULSM not only highly accurate but also computationally viable for industrial-level traffic under stringent latency constraints (details in Appendix \ref{appendix:a}).

\subsection{Ablation Study}

\begin{table}[htbp]
\centering
\small
\caption{UxSID ablation results. ‘-’ indicates ablation-induced or feature unavailability.}
\label{tab:combined-ablation-study}
\setlength{\tabcolsep}{2pt} 
\begin{tabular}{lcccccccccc}
\toprule
\multirow{3}{*}{Variants} & \multicolumn{7}{c}{Industrial Dataset} & \multicolumn{3}{c}{Public Datasets} \\
\cmidrule(lr){2-8} \cmidrule(lr){9-11}
& \multicolumn{3}{c}{CTR} & \multicolumn{3}{c}{CTCVR} &  & XLong & \multicolumn{2}{c}{KuaiRec-Big} \\
\cmidrule(lr){2-4} \cmidrule(lr){5-7} \cmidrule(lr){9-9} \cmidrule(lr){10-11}
& AUC & UAUC & WUAUC & AUC & UAUC & WUAUC & Int.R@50 & AUC & AUC & Int.R@50 \\
\midrule
Category (Tag) & 0.8707 & 0.6081 & 0.6088 & 0.8605 & 0.6186 & 0.6288 & 0.0543 & - & 0.8261 & 0.0916 \\
\midrule
w/o $\mathbf{e}_{global}$ & 0.8714 & 0.6101 & 0.6108 & 0.8615 & 0.6230 & 0.6318 & - & 0.8370 & 0.8302 & - \\
w/o $\mathbf{e}_{local}$  & 0.8719 & 0.6114 & 0.6121 & 0.8618 & 0.6238 & 0.6327 & 0.1454 & 0.8375 & 0.8314 & 0.2009 \\
w/o $\mathcal{L}_{ortho}$ & 0.8725 & 0.6119 & 0.6144 & 0.8624 & 0.6261 & 0.6342 & 0.1471 & 0.8385 & 0.8344 & 0.2063 \\
w/o Gate                  & 0.8723 & 0.6116 & 0.6127 & 0.8623 & 0.6249 & 0.6334 & 0.1467 & 0.8376 & 0.8342 & 0.2044 \\
\midrule
\textbf{UxSID}     & \textbf{0.8728} & \textbf{0.6125} & \textbf{0.6161} & \textbf{0.8626} & \textbf{0.6269} & \textbf{0.6350} & \textbf{0.1488} & \textbf{0.8408} & \textbf{0.8348} & \textbf{0.2071} \\
\bottomrule
\end{tabular}
\end{table}

We perform a comprehensive ablation study on UxSID to assess its components and target semantic-specific perception (Table \ref{tab:combined-ablation-study}). 

\textbf{Effectiveness of SID-based Semantic Querying.} We evaluate query granularity by replacing candidate SIDs with coarse-grained category (tag) attributes. As shown in Table \ref{tab:combined-ablation-study}, SIDs consistently outperform tag-based queries across all metrics. This confirms that category-level attributes lack the semantic resolution required to precisely navigate complex user interest distributions.

Conversely, SIDs act as high-density semantic probes, enabling granular activation of historical interests. As evidenced by the Int.R@50 metric, this probing mechanism enhances the model’s ability to identify historical interactions within the target's semantic cluster. This increased recall likely drives the observed AUC gains, empirically demonstrating that UxSID successfully bridges the target-agnostic bottleneck inherent in traditional compression paradigms.

\textbf{Impact of Hierarchical Semantic Probing.} Removing either $\mathbf{e}_{global}$ or $\mathbf{e}_{local}$ leads to noticeable performance degradation. Specifically:
\begin{itemize}[leftmargin=*]
    \item \textbf{Compressed Global Signals (\textit{w/o $\mathbf{e}_{global}$})}: Eliminating the explicit attention results in a significant drop in AUC. This indicates that fine-grained, item-to-item semantic signals are indispensable for capturing immediate target-specific relevance that might be smoothed over during compression.
    \item \textbf{Compressed Local Interests (\textit{w/o $\mathbf{e}_{local}$})}: The observed decline suggests that interest anchors are crucial for filtering historical noise and providing a diverse, structural view of user preferences that $\mathbf{e}_{global}$ cannot effectively capture.
\end{itemize}

\textbf{Role of Gated Probing and Diversity Loss.} Ablating $\text{GatedNet}(\cdot)$ reveals that direct probing with original SIDs is suboptimal. The gating mechanism ensures the local latent probe is precisely conditioned on the global context from the first stage, enhancing robustness during querying. Furthermore, removing $\mathcal{L}_{ortho}$ adversely affects performance, validating that forcing distinct interest clusters prevents mode collapse and preserves the multi-faceted nature of user preferences.

Appendix \ref{app:Performance Gain Attribution} confirms UxSID's gains stem from its architecture rather than SIDs injection.

\subsection{Efficiency and Scaling Law Analysis}
\label{sec:efficiency_scaling}

Theoretical complexities are summarized in Table~\ref{tab:complexity}. Search-based paradigms (SIM, TWIN) suffer from online matching overhead. Conversely, UxSID offloads ULSM offline, achieving $O(1)$ online complexity for target-level compression and maintaining constant latency even when scaling to 10k.

\begin{wraptable}{R}{0.5\textwidth}
  \small
  \centering
  \caption{Inference time complexity comparison. $B$: batch size, $L$: original sequence length, $R$: retrieved sequence length, $d$: hidden size, $A$: attribute index size, $m$: hash functions, $c$: compressed interest length, $f$: feature numbers.}
  \label{tab:complexity}
  \begin{tabular}{lc}
    \toprule
    \textbf{Model} & \textbf{Inference Time Complexity} \\
    \midrule
    SIM-Hard & $(B \log(A) + BRd)$ \\
    SIM-Soft & $(BLd + BRd)$ \\
    ETA      & $(BLm + BRd)$ \\
    SDIM     & $Bm\log(d)$               \\
    MIRRN    & $BLm + BR\log(R)d + BRd^2$ \\
    TWIN     & $(BL + BfLd + BRd)$ \\
    C-Former & $(BRd)$ \\
    \textbf{UxSID } & $(Bcd)$ \\
    \bottomrule
  \end{tabular}
\end{wraptable}

\begin{figure}[t]
  \centering
  \includegraphics[width=1\linewidth]{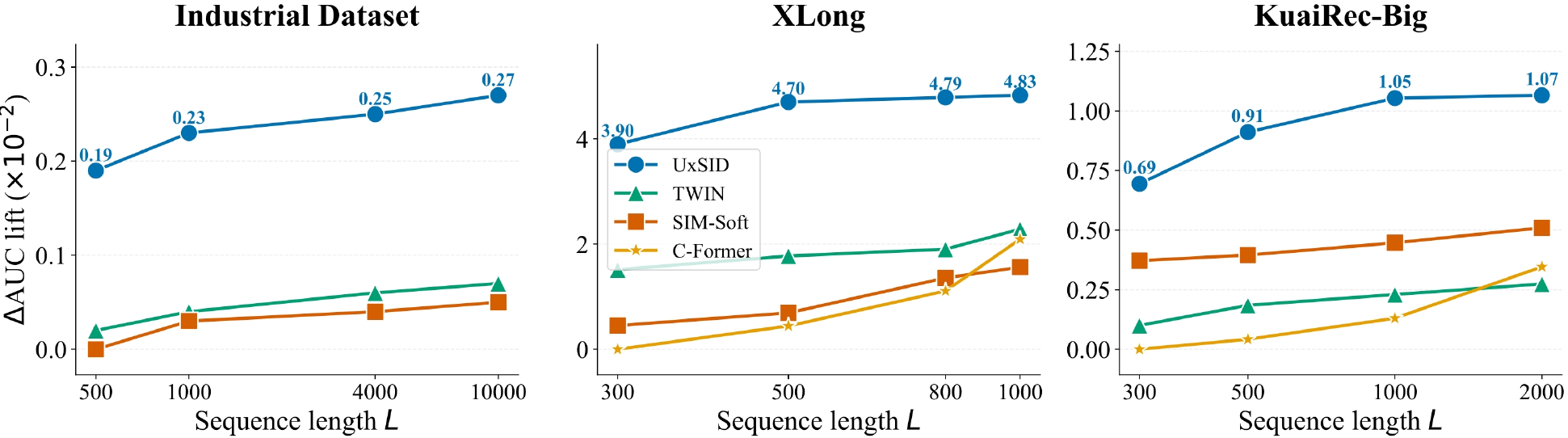}
  \caption{AUC improvements (percentage points) across various sequence lengths on all datasets.}
  \label{fig:scaling}
\end{figure}

Figure~\ref{fig:scaling} illustrates the performance trends as sequence length scales, revealing two primary findings:

\textbf{Limitations of Retrieval and Static Compression.} As sequence lengths increase, search-based models (SIM and TWIN) exhibit decelerating growth; their fixed retrieval scope inevitably excludes distal but relevant interactions. Simultaneously, C-Former’s performance is unstable on short sequences but scales as histories lengthen. This highlights a flaw in static compression: without target-aware guidance, the model struggles to disentangle signals from mounting noise, failing to maintain robust representations across sequence scales.

\textbf{Scalability and Efficiency.} UxSID consistently maintains the highest AUC, with the performance gap widening at the 10k scale. This confirms that SID-based routing effectively pinpoints specific interests, translating expansive behavioral data into substantial gains. Notably, this is achieved with constant online consumption, demonstrating UxSID's potential for modeling lifelong behaviors.

%
%
\subsection{Storage Feasibility Analysis}
\label{sec:storage}

Deployability of this lookup depends on how many keys the cache holds.
User-level compression stores one group of tokens per user, while UxSID stores one compressed representation per active SID of that user and thus has a larger key space in form.
The key set stays small in practice, because a user's interests concentrate in a limited number of SID buckets.
Each cached value is the representation of shape \([2,32]\), at \(s=128\) bytes per key.
On a platform with \(|U|=4\times10^{8}\) daily active users and about \(50\) active SIDs per user monthly, the cache holds \(N_{\mathrm{keys}}\approx 2\times10^{10}\) keys and occupies \(\mathrm{Mem}\approx 2.5~\text{TB}\).
A few terabytes lie within the capacity of a distributed key-value store, so Eq.~(12) is a constant-time read over a bounded key set.
The same count is capped by the codebook: \(n_{\mathrm{sid}}\) tracks how many semantic buckets a user has touched, and a user's interests are finite, so the working set stays near this scale however long the history grows. An item-specific cache has no such cap.
Its size is \(|U|\) times the number of distinct videos each user has watched, and it grows linearly with that viewing count.
At the traffic of large industrial platforms, the resulting volume exceeds the petabyte scale, varying with the scenario, and it keeps rising as users watch more videos.
UxSID does not have this problem in real storage.

Cache refresh and sequence length leave this storage budget unchanged, and it grows only when the user watches a video whose SID the user has not seen.
An already-seen SID overwrites the same entry and allocates no new key, so existing keys stay unchanged.
History length behaves the same way.
The number of compressed tokens is a fixed hyper-parameter: users with different sequence lengths produce interest representations of the same size, so both the storage size and the online cost are invariant to \(L\).
Only the offline compression compute grows with \(L\).
The longer histories in Figure \ref{fig:scaling} are therefore absorbed at no extra storage and no extra serving cost.

\subsection{Parameter Sensitivity Study}
\label{sec:parameter_study}

\begin{figure}[t]
  \centering
  \includegraphics[width=1\linewidth]
  {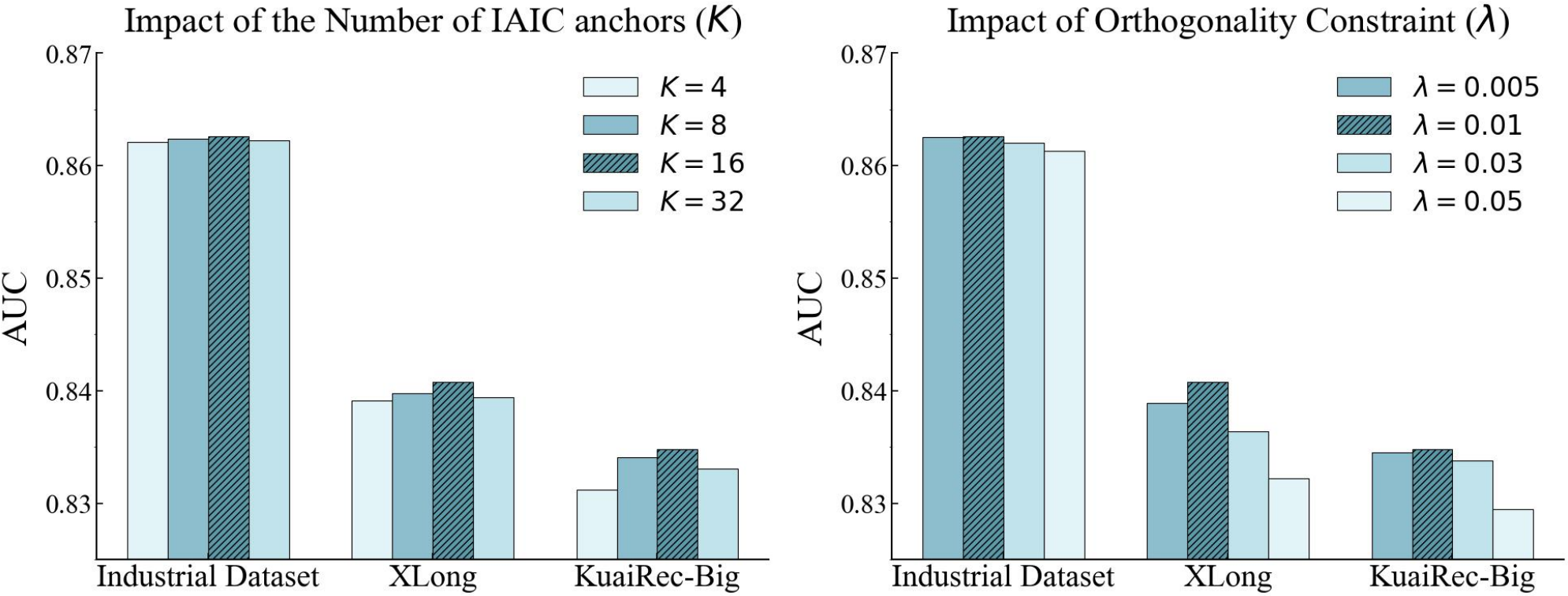}
  \caption{Hyperparameters analysis of UxSID on all three datasets.}
  \label{fig:params analysis}
\end{figure}

We conduct a sensitivity analysis across all datasets, with results summarized in Figure \ref{fig:params analysis}.

\textbf{Impact of the Number of IAIC anchors ($K$).}
$K$ dictates the capacity of the IAIC block to capture diverse intents. At $K=4$, performance drops, suggesting that compressing ultra-long behavior into a highly bottlenecked embedding forces interest entanglement and the loss of fine-grained information. Performance improves as $K$ grows, peaking at $16$. However, excessively large $K$ introduces redundancy and noise, causing overly dispersed routing and slightly degraded metrics.

\textbf{Impact of Orthogonality Constraint ($\lambda$).}
$\lambda$ controls the intensity of the diversity regularization applied to the compressed interests. Without sufficient penalty, the interest anchors tend to capture high-frequency patterns, reducing the diversity of the IAIC block. However, an excessively large $\lambda$ overly constrains the latent space, forcing separation that disrupts the primary CTR prediction.

\subsection{Visualization and Case Study}

\begin{figure}[t]
  \centering
  \includegraphics[width=1\linewidth]{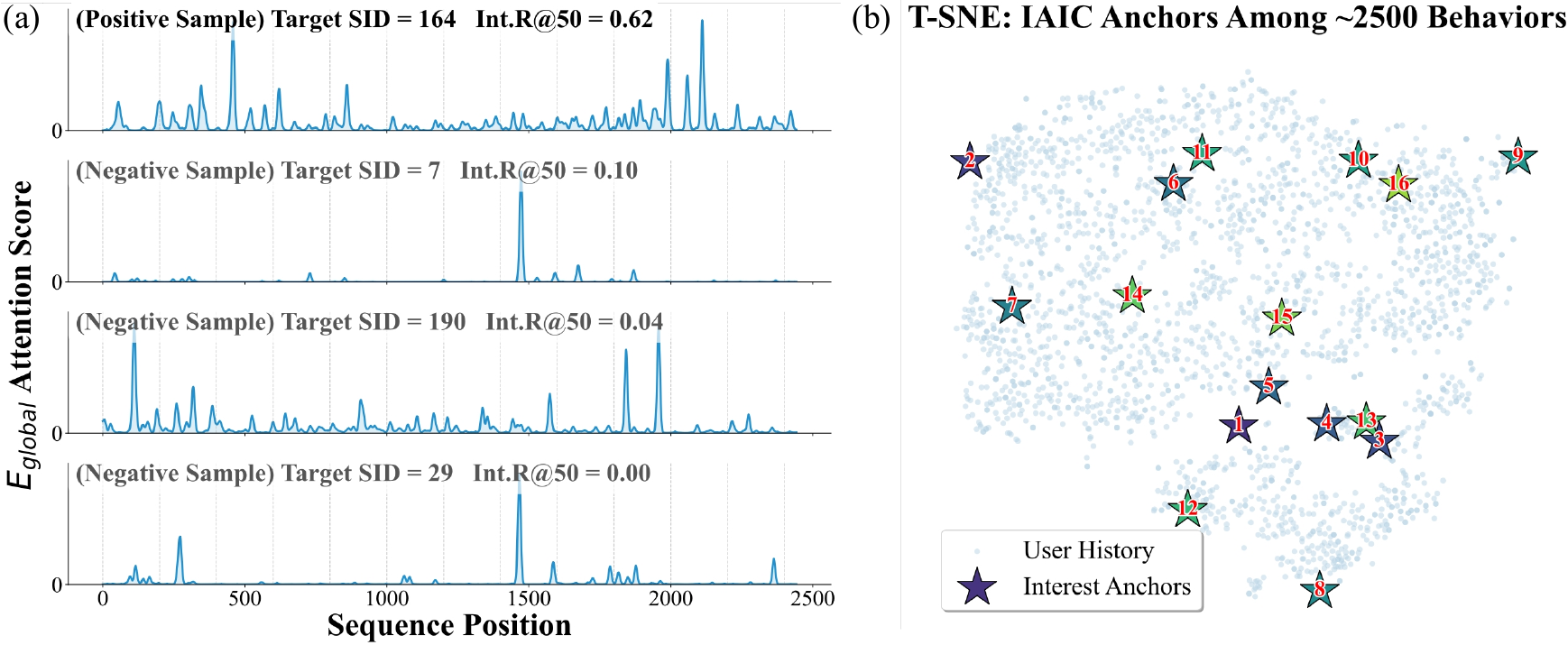}
  \caption{Efficacy of UxSID in interest modeling. (a) highlights the target SID-based attention routing, while (b) showcases the diversity of learned long-term interest anchors.}
  \label{fig:case}
\end{figure}

We investigated the mechanisms underlying UxSID’s efficacy, focusing on candidate SID activation within sequences and compressed anchors on KuaiRec-Big. As shown in Figure \ref{fig:case}(a), distinct attention routing patterns for different candidates demonstrate that SIDs successfully trigger diverse semantic interests. The contrast in Int.R@50 between positive and negative samples validates the alignment between activation relevance and CTR objectives: negative items yield low Int.R@50 due to semantic irrelevance, whereas positive samples achieve significantly higher recall. Notably, UxSID activates data across the entire behavioral cycle; even early behaviors are assigned high attention weights, a granularity traditional compression methods struggle to achieve. Figure \ref{fig:case}(b) further reveals that the IAIC block captures multiple interest anchors, covering the full spectrum of behaviors. This synergy ensures both the efficiency and comprehensiveness of UxSID in ULSM.

%% file: Conclusion.tex
\section{Conclusion}
\label{Conclusion}
In this paper, we propose UxSID, an end-to-end framework that leverages target SIDs for compressed ultra-long sequence modeling in modern recommender systems. UxSID introduces the Item-Agnostic Interest Compression mechanism and Hierarchical Semantic Probing, which operationalize SIDs as semantic-specific routers to navigate complex, ultra-long behavioral histories. By harnessing the high-density semantic resolution of SIDs, the framework effectively bridges the gap between historical interaction density and target intent precision, achieving superior predictive accuracy while maintaining industrial-level efficiency. Extensive offline and online experiments demonstrate that UxSID significantly outperforms SOTA baselines, while comprehensive ablation studies validate the necessity and effectiveness of each constituent module. Future work involves the full-scale deployment in Kuaishou’s short-video scenarios and its extension to lifelong sequential modeling.

\begin{ack}
This work was fully supported by Kuaishou Technology.
\end{ack}

%% file: Appendix.tex
\appendix
\newpage

\section{Related Work}
\label{Related Work}

\textbf{Foundations of Sequence Modeling and the Scalability Bottleneck.} Historically, industrial recommendation systems have adhered to a paradigm that combines sequence modeling with feature interaction. A defining milestone in this trajectory is DIN \cite{din}, which introduced a target-aware attention mechanism to dynamically activate historical behaviors relevant to the candidate item. Subsequent innovations expanded on this foundation to capture temporal dependencies (e.g., DIEN \cite{dien} and DSIN \cite{dsin}), mitigate noise within sequences, and extract multi-faceted user interests \cite{chai2022user,li2019multi,M-GPT,bst}. With the success of Transformer \cite{attention}, self-attentive architectures such as SASRec \cite{sasrec} have become standard practice. More recently, generative models optimized via next-token prediction, such as HSTU \cite{hstu} and MTGR \cite{mtgr}, have emerged, demonstrating formidable capabilities in learning rich, contextualized user interest representations. However, the quadratic computational complexity inherent in these attentive and generative architectures imposes a severe sequence-length scalability challenge, strictly bounding their deployable length within the stringent millisecond-level latency constraints of online serving environments \cite{ULRM}. This bottleneck inherently truncates the model's receptive field, rendering them inadequate for capturing long-term, heterogeneous user trajectories, thereby necessitating new works specifically tailored for ULSM \cite{SRsurvey1}.

\textbf{Search-Based Paradigms for Ultra-Long Sequence Modeling.} To overcome the scalability bottleneck of ultra-long sequences, the industry gravitates toward a two-stage cascaded framework—conceptualized as the ``Global Search First, Exact Search Later'' paradigm—wherein a lightweight GSU filters the expansive history into a target-relevant top-$K$ sub-trajectory for subsequent fine-grained attention modeling by a more expressive Exact Search Unit (ESU). Early approaches, such as SIM \cite{SIM}, relied on hard filtering based on categorical item attributes, whereas subsequent models introduced hashing and low-precision attention (e.g., ETA \cite{eta} and SDIM \cite{sdim}) to accelerate similarity computation. More recently, TWIN \cite{twin} proposed an end-to-end architecture to align the optimization objectives of both stages, a framework that was subsequently enhanced in TWINv2 \cite{twinv2} via hierarchical clustering to further boost scalability and performance. 

Despite their success, search-based methods suffer from an inherent limitation: their filtering logic often sacrifices historical items with low superficial correlation but high latent semantic synergy \cite{MIRRN}. For example, a search unit biased toward "pants" may discard a matching belt or stylistic footwear, missing crucial combinatorial intents. In contrast, SIDs transcend these surface-level filters, capturing high-order semantic links often invisible to traditional retrieval \cite{tiger,onemall}.



\textbf{Compression-Based Paradigms for Ultra-Long Sequence Modeling.} To circumvent the prohibitive online computational costs of search-based methods and better exploit the global contextual information inherent in long sequences, compression-based approaches \cite{mimn,lurm,pinnerformer,hpmn,lrea,dv365,longer,marm} have emerged. These methods primarily aim to offload the heavy burden of sequence modeling to the offline stage, compressing ultra-long sequences into compact, reusable user representations. Early explorations include memory-augmented models like MIMN \cite{mimn} and HPMN \cite{hpmn}, which leverage memory networks to effectively compress sequences at the scale of thousands of interactions offline.
Recent advancements focus on modularizing and scaling this offline representation learning. For instance, LREA \cite{lrea} employs low-rank matrix factorization to compress sequence representations, enabling downstream models to efficiently capture user interests via target attention over these condensed embeddings. Alternatively, rule-based partitioning methods, such as DV365 \cite{dv365}, slices sequences into fixed-length or temporal windows. PinnerFormer \cite{pinnerformer}, LURM \cite{lurm}, and C-Former \cite{C-former} condense expansive user sequences into interest clusters or dense embeddings.

While compression-based methods excel in online efficiency, static compression inherently acts as a low-pass filter—retaining coarse-grained global trends while marginalizing the high-frequency, target-specific interest peaks critical for precise ranking. This limitation underscores a pressing need for adaptive compression frameworks capable of generating target semantic-aware representations.

\textbf{Item Tokenization and Semantic Identifiers.} Item identifiers are fundamental to modern recommender systems. Current paradigms are transitioning from random Photo IDs (PIDs) \cite{din,SIM}, which rely on embedding lookups, toward SIDs constructed via RQ-VAE \cite{tiger} or other tokenization frameworks (e.g., QARM \cite{luo2025qarm}). Distinct from existing paradigms that primarily utilize SIDs for generative recommendation \cite{onerec,plum,pit} or item representation enrichment \cite{qarmv2,farewell}, we propose to employ SIDs as dynamic semantic queries within the ULSM context. Their clustered semantic codes enable navigating complex interest spaces where traditional PID-based models lack sufficient granularity.

\section{Online Implementation}
\label{appendix:a}

\begin{figure}[htbp]
  \centering
  \includegraphics[width=\linewidth]{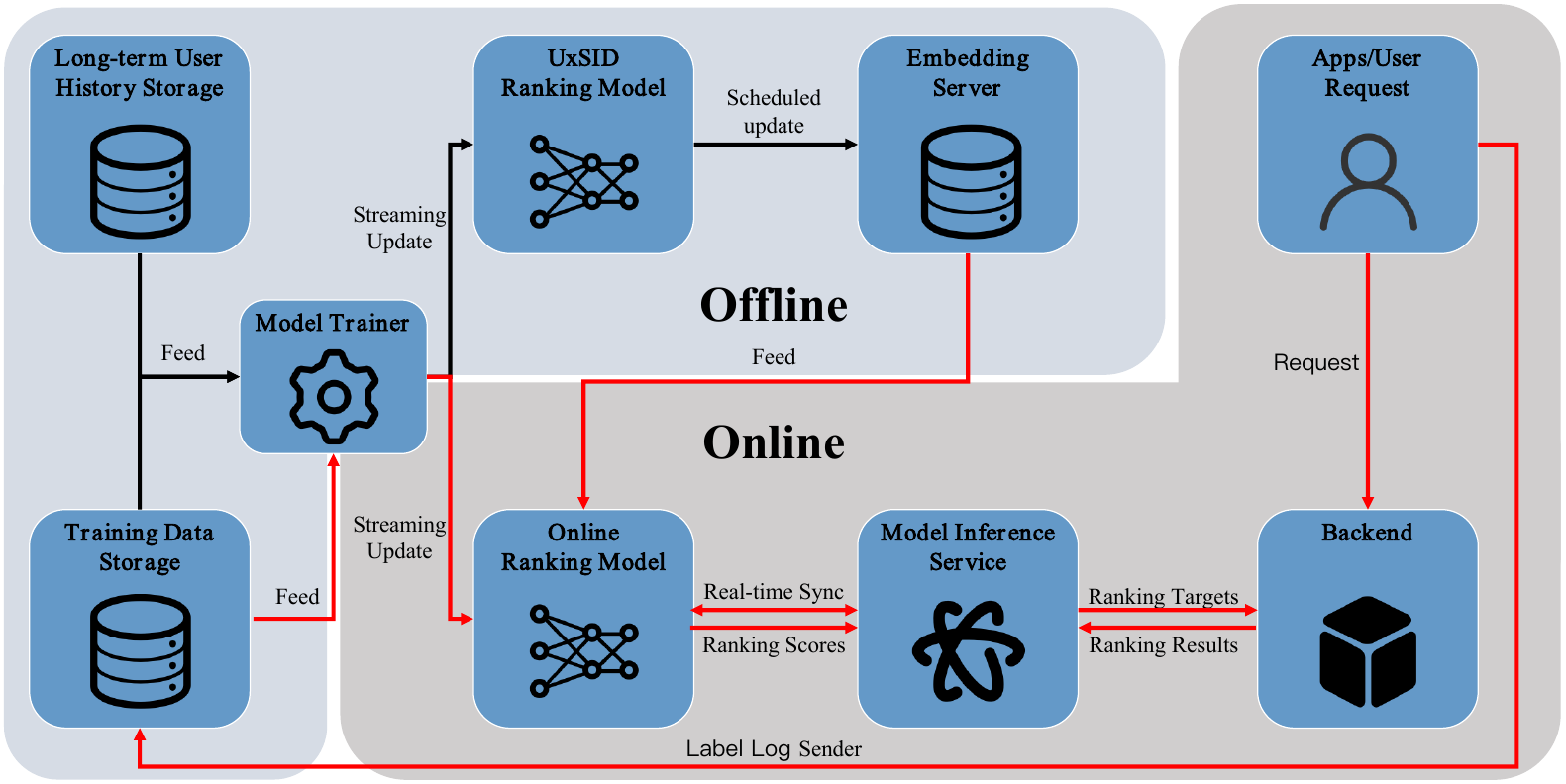}
  \caption{The overall system deployment pipeline of UxSID, comprising offline UxSID embedding generation (black path), online model training (red path), and real-time online inference (red path).}
  \label{fig:pipeline}
\end{figure}


\textbf{Model Configurations and Maintenance.} UxSID utilizes the same SIDs codebook architecture as QARM V2 \cite{qarmv2} and OneRec-v2 \cite{zhou2025onerecv2}
, which has demonstrated outstanding performance across multiple industrial scenarios. Specifically, the first-level codebook size is set to 4096, with an embedding lookup dimension of 32. This dimension is strategically aligned with the hidden units of the three single-head attention networks within UxSID to ensure representation consistency. To maintain the timeliness of the semantic representations, the embedding server undergoes a scheduled update once per week. The comprehensive end-to-end pipeline, encompassing both offline generation and online serving, is illustrated in Figure \ref{fig:pipeline}.

\textbf{Computational Cost and Latency Analysis.} We evaluate the resource efficiency of UxSID through both training and inference stages. For the UxSID model offline training, a sequence length of 1k requires 16 GPUs, while scaling to a 10k sequence length necessitates 40 GPUs. Crucially, the resource consumption of the downstream online model remains invariant to the sequence length, as it only interacts with the compressed UxSID embeddings (of shape $[2, 32]$) in an incremental manner. In terms of serving infrastructure, while the current online model with a 1k sequence length occupies approximately 450 GPUs, we estimate that directly scaling the sequence length to 10k within the ranking model would necessitate approximately 5,300 GPUs to meet throughput requirements. By shifting the computational burden of long-sequence modeling to the UxSID framework, our approach introduces a negligible latency overhead of only +0.16 ms compared to the baseline without UxSID embeddings. This balance between high-capacity modeling and minimal inference latency underscores the efficiency of UxSID for real-time recommendation at scale.

\section{Datasets}
\label{app:datasets}
\begin{table}[htbp]
\centering
\small
\caption{Statistics of public datasets.}
\label{tab:dataset}

\begin{tabular}{lrrrrr}
\toprule
Dataset & \#Users & \#Items & \#Interaction & \begin{tabular}[c]{@{}c@{}}Avg. Seq.\\Length\end{tabular} & \begin{tabular}[c]{@{}c@{}}Max. Seq.\\Length\end{tabular} \\
\midrule
XLong & 1,000 & 3,269,017 & 1,000,000 & 1,000 & 1,000 \\
KuaiRec-Big & 7,176 & 10,728 & 12,530,806 & 1,746 & 3,000\\

\bottomrule
\end{tabular}
\end{table}
\textbf{XLong \footnote{https://tianchi.aliyun.com/dataset/22482} \cite{XLong}:} Sampled from a large e-commerce platform (April–September 2018), this dataset provides sequences of about 1k interactions per user for long-term modeling. For each item sequence, the last video is used for testing, the video before the last is used for validation, and the rest is used for training.\\
\textbf{KuaiRec-Big \footnote{https://kuairec.com/} \cite{kuairec}:} KuaiRec is a real-world dataset collected from the recommendation logs of the app Kuaishou, containing user interactions and video metadata from July 5, 2020 to September 5, 2020.  We use users’ interaction histories to create video sequences sorted by timestamp and filter out users with less than 50 comments. For each video sequence, the last 7 videos are used for testing, the 8th to 14th last videos are used for validation, and the rest are used for training. We fix the user's interaction history up to 2k.\\
\textbf{Industrial Dataset:} Collected from Kuaishou's advertising system (April 1-7 2026), this dataset contains impression logs and labels. For each user, the interaction history is preserved up to 10k.\\
\textbf{SIDs Generation.} For the KuaiRec-Big dataset, we extract item-related fields from item\_daily\_features.csv and employ the Llama model \cite{touvron2023llama} to generate high-dimensional content embeddings. The XLong dataset comes with pre-provided item embeddings, which we utilize directly. For both datasets, we apply the quantization toolkit provided by LETTER \footnote{https://github.com/HonghuiBao2000/LETTER/tree/master}\cite{letter} to train the hierarchical codebooks, resulting in SIDs with a structured configuration of [$256 \times 256 \times 256 \times 256$]. 

\section{Baselines}

Below we briefly describe baselines with which we compare UxSID:
\label{app:baselines}
\begin{itemize}[leftmargin=*]
\item \textbf{DIN \cite{din}} is a foundational baseline that introduces a target-aware attention mechanism from recent behavior sequences.
\item \textbf{SIM Hard / Soft} \cite{SIM} adopts a two-stage approach where a GSU filters the expansive history via either category-based hard matching or embedding-based soft retrieval.
\item \textbf{ETA} \cite{eta} employs Locality Sensitive Hashing and Hamming distance for GSU retrieval. 
\item \textbf{SDIM} \cite{sdim} adopts a sampling-based approach that utilizes multiple hash functions to generate signatures for ULSM.
\item \textbf{MIRRN} \cite{MIRRN} utilizes multi-granularity queries across diverse time scales for behavior retrieval.
\item \textbf{TWIN} \cite{twin} ensures target consistency between the GSU and ESU stages for ULSM.
\item \textbf{C-Former} \cite{C-former} employs reconstruction constraints to cluster user interests in an end-to-end manner.
\end{itemize}

\section{Detailed Results on Public Datasets}

\subsection{Robustness and Stability Analysis}
\label{app:Stability Analysis}
To evaluate the stability of UxSID, we report the performance across three independent runs using different random seeds (2024, 2025, and 2026). As shown in Table \ref{tab:detailed_seed_results}, the marginal variance observed across these runs demonstrates that our framework is highly robust against initialization noise and consistently maintains its performance gains across different experimental trials.

\begin{table}[htbp]
\centering
\small
\caption{Detailed performance of UxSID across three independent runs with different random seeds (2024, 2025, 2026). The final results reflect the stability of our proposed framework.}
\label{tab:detailed_seed_results}
\begin{tabular}{lccc}
\toprule
\textbf{Model} & \textbf{Seed} & \textbf{KuaiRec-Big (AUC)} & \textbf{XLong (AUC)} \\
\midrule
UxSID & 2024 & 0.83484 & 0.84084 \\
UxSID & 2025 & 0.83482 & 0.84405 \\
UxSID & 2026 & 0.83462 & 0.83964 \\
\midrule
\textbf{Avg $\pm$ Std} & - & \textbf{0.8348 $\pm$ 0.0001} & \textbf{0.8415 $\pm$ 0.0023} \\
\bottomrule
\end{tabular}
\end{table}

\subsection{Performance Gain Attribution}
\label{app:Performance Gain Attribution}
A potential concern is whether the performance gain of UxSID stems primarily from the item-side SID information rather than the proposed architecture. To isolate the impact of this information enrichment, we conduct a controlled experiment where the first-layer SID is incorporated as an additional sparse feature for all baseline models.

Table \ref{tab:sid_augmented} shows that while the inclusion of SID features slightly improves the performance of all baselines, UxSID consistently maintains a substantial lead. This confirms that our superiority is fundamentally derived from the structural design of the IAIC and the hierarchical routing mechanism, rather than mere information enrichment.

\begin{table}[htbp]
\centering
\small
\caption{Comparison with baselines after augmenting all models with item-side SID features. The best results are in bold.}
\label{tab:sid_augmented}
\begin{tabular}{lcc}
\toprule
\textbf{Models (+ SID Features)} & \textbf{XLong} & \textbf{KuaiRec-Big} \\
\midrule
DIN          & 0.7932          & 0.8214          \\
SIM-Hard     & --              & 0.8246          \\
SIM-Soft     & 0.7999          & 0.8305          \\
ETA          & 0.7930          & 0.8271          \\
SDIM         & 0.7955          & 0.8246          \\
MIRRN        & 0.7976          & 0.8257          \\
TWIN         & 0.8189          & 0.8296          \\
C-Former    & 0.8180          & 0.8311          \\
\midrule
UxSID (Base) & 0.8408 & 0.8348 \\
\textbf{UxSID + SID} & \textbf{0.8439} & \textbf{0.8361} \\
\bottomrule
\end{tabular}
\end{table}

\section{Limitation}
\label{app:limitation}
Despite its efficiency, the industrial deployment of UxSID presents certain specialized requirements that could be challenging in general settings. First, the framework's performance is intrinsically linked to the quality of pretrained SIDs; in massive-scale production environments, ensuring the real-time availability of SIDs for cold-start items (i.e., fetching SIDs during the very first user request) requires a highly optimized streaming inference pipeline. Second, maintaining the large-scale storage and high-concurrency retrieval efficiency of SID embeddings poses non-trivial engineering hurdles. These remain key considerations for adapting UxSID to other large-scale recommendation systems.